# Evaluation of Rounding Functions in Nearest-Neighbor Interpolation

Olivier Rukundo
Norwegian Colour and Visual Computing Laboratory, Department of Computer Science,
Norwegian University of Science and Technology, Gjøvik, Norway

## ABSTRACT

A novel evaluation study of the most appropriate round function for nearest-neighbor (NN) image interpolation is presented. Evaluated rounding functions are selected among the five rounding rules defined by the Institute of Electrical and Electronics Engineers (IEEE) 754-2008 standard. Both full- and no-reference image quality assessment (IQA) metrics are used to study and evaluate the influence of rounding functions on NN interpolation image quality. The concept of achieved occurrences over targeted occurrences is used to determine the percentage of achieved occurrences based on the number of test images used. Inferential statistical analysis is applied to deduce from a small number of images and draw a conclusion of the behavior of each rounding function on a bigger number of images. Under the normal distribution and at the level of confidence equals to 95%, the maximum and minimum achievable occurrences by each evaluated rounding function are both provided based on the inferential analysis-based experiments.

## 1. INTRODUCTION

Interpolation is a widely used method, in many fields, to construct a new data value within the range of a set of known data[1-6, 23]. In the video and/or dynamic imaging, if the interpolation method becomes too computationally inefficient or time-consuming, it may lead to the jerky appearance of images. In image upscaling or high-resolution or resolution enhancement, if the interpolation method is not accurate enough, it may result in heavy error propagation or visual artefacts, particularly at the edges of upscaled image objects. Visual artefacts - such as aliasing/jaggy, blurring, and edge-halo artefacts - are important contributors to the loss of image interpolation quality. However, the image interpolation quality can also decrease as the scaling ratio increases significantly, which ideally should not have to be the case. In digital zoom, one of the advantages of image upscaling is the possibility to get a closer view of objects in small-sized images and/or videos[7-10], without the need for a mechanical device of lens elements such as the one used in optical zoom. Many works on image interpolation reported new strategies for the minimization of visual artefacts at specific scaling ratios. Those strategies can be classified into adaptive[11-13] and non-adaptive[14-16] as well as, very recently, into non-extra pixel and extra pixel categories[21]. Given that the nearest-neighbor algorithm performance depends entirely on the accuracy or precision of the rounding function, with such a computational simplicity, this algorithm has become the fastest and crispest image-edge productive among other/existing image interpolation approaches and algorithms[3,5,15]. The disadvantage of using the nearest-neighbor algorithm for image interpolation is the production of upscaled images with the most jagged-edges among other well-known algorithms[5]. Such a disadvantage is mainly linked to the loss of precision of a given rounding function used to round-off any scaled-coordinates of non-integer type - due to the linear scaling equation and need for integer usage[21]. There exist many rounding functions and rules which can round-off a non-integer output to an integer output. The IEEE 754-2008 standard defined five rounding rules[17]. Here, rounding-off or rounding a non-integer means transforming some non-integer quantity from a greater precision to lesser precision[22]. In NN image interpolation, such a lesser precision has a direct effect on which pixel to pick from the source image and copy in the destination image. In the attempt to reduce the imprecision effects on NN image interpolation quality, it is important to understand the influence of rounding functions on the NN image interpolation algorithm to determine the rounding function leading to the least error propagation damages. In this paper, a novel study is presented to answer that question. The paper is organized as follows: Part I introduces the paper. Part II recaps the nearest-neighbor algorithm and rounding rules. Part III presents experimental evaluations. The evaluation conclusion is given in Part IV.

## 2. THE NEAREST NEIGHBOR AND ROUNDING FUNCTIONS

The linear scaling equation on which the nearest-neighbor interpolation algorithm is based allows scaling a given image to a desired or new image size. This can be achieved thanks to two important mathematical operations - namely *rounding* and *linear scaling* - on which runs the nearest-neighbor algorithm. Eq.1 gives the linear scaling equation and shows four elements involved in the scaling operations; where $srcLength$ is a variable representing the length of the source image, $dstCoord$ is a variable representing the destination coordinates, $scrCoord$ is a variable representing the source coordinates, and $destLength$ is a variable representing the length of the destination image.

$$\frac{scrCoord}{scrLength} = \frac{destCoord}{destLength} \tag{1}$$

The ratio between the destination and source length variables (i.e. $destLength / srcLength$) is a constant equivalent to the scaling factor or ratio (***r***). The source and destination coordinates are conventionally expected to be of integer type quantities.

Table 1: Five rounding rules defined by the IEEE 754-2008 standard

| **RULES** | **EXAMPLES OF HALF-INTEGERS** | | | |
|---|---|---|---|---|
| | +11.5 | +12.5 | -11.5 | -12.5 |
| round to nearest, ties/half-integers to even | +12.0 | +12.0 | -12.0 | -12.0 |
| round to nearest, ties/half-integers away from zero (*round*) | +12.0 | +13.0 | -12.0 | -13.0 |
| Round toward 0 (*fix*) | +11.0 | +12.0 | -11.0 | -12.0 |
| round toward $+\infty$ (*ceil*) | -12.0 | -13.0 | -11.0 | -12.0 |
| round toward $-\infty$ (*floor*) | +11.0 | +12.0 | -12.0 | -13.0 |

To ensure this is respected, when Eq. 1 gives coordinates of non-integer type, rounding operation is required and must be performed to meet the digital format requirement. Table 1 shows five rounding rules defined by the IEEE 754-2008 standard as well as Maxfield's diagram[17, 22].

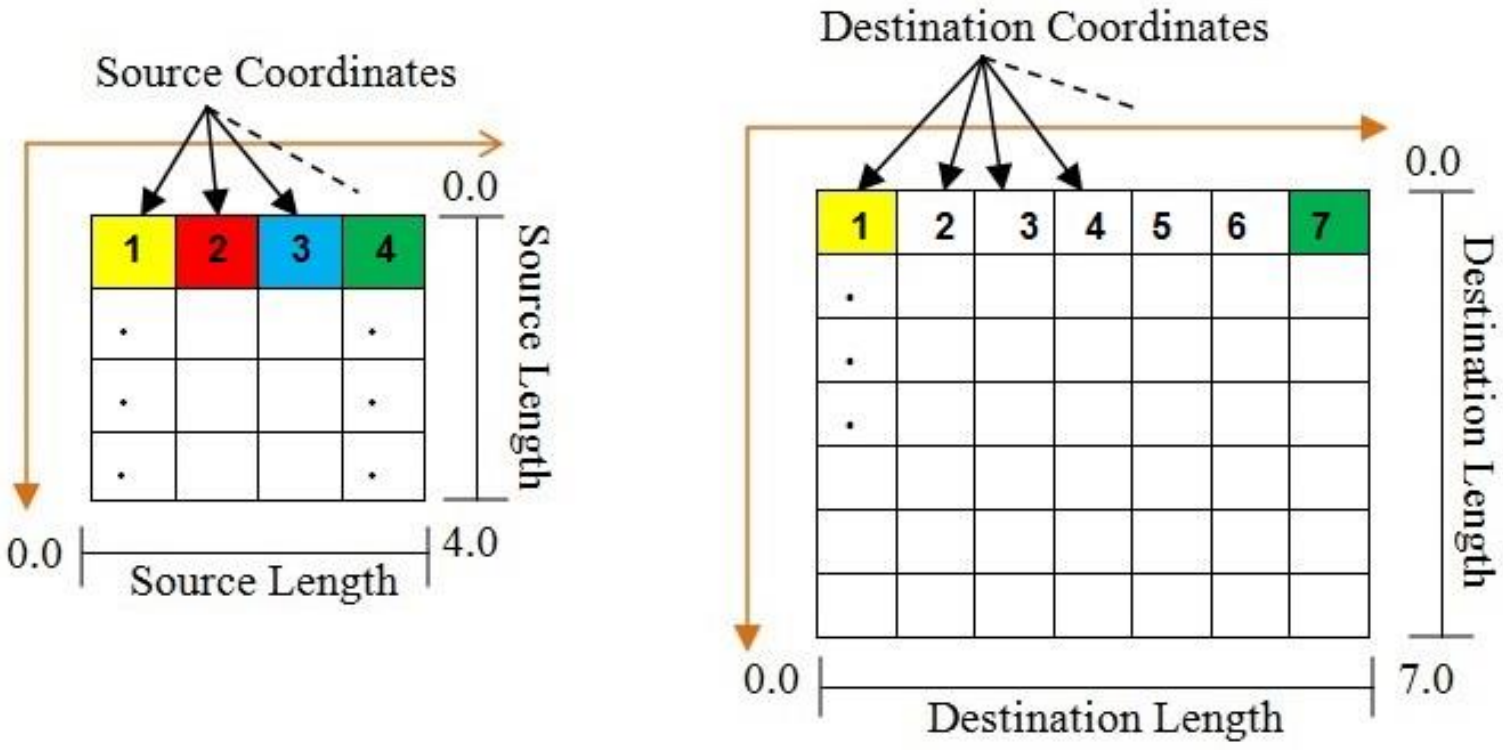

Figure 1: Example of source and destination images coordinates

However, the scope of this work encompasses three of five namely *floor*, *ceil,* and *round* with the main objective to evaluate their effect on image interpolation quality. As can be seen in Table 1, *floor* means rounding towards minus infinity, *ceil* means rounding towards plus infinity, and *round* means rounding to the nearest integer, if a non-integer has a tie or if it is a half-integer round to the nearest integer away from zero. Figure 1 shows the example of the source and destination images with their coordinates and lengths added. As can be seen, the source image has four coordinates or indices, namely 1, 2, 3, 4, and its length equals four. The destination image has seven coordinates, namely 1, 2, 3, 4, 5, 6, 7, and its length equal to seven. Since the source and destination images are not equal in lengths, the source image pixels are insufficient to fill in or complete the destination image. It is, therefore, necessary to use the linear scaling equation to find all coordinates correspondences to finally be able to approximate the missing pixels and fill in all pixel locations in the destination image. Unlike other interpolation algorithms, the nearest-neighbor algorithm does not create extra-pixels to find additional pixels to use during image upscaling[21]. Extra-pixels are pixels that do not belong to the source image[21]. Table 2 contains information demonstrating the nearest-neighbor algorithm's strategy of finding the missing pixels. As can be seen, the first column represents the destination coordinates shown in Figure 1. The second column represents the Eq.1 with its variables shown in Figure 1. The third column contains linearly calculated source coordinates thanks to Eq.1. The fourth, fifth, and sixth columns show the integers achieved from using the *floor*, *ceil*, and *round* functions, respectively. Now, to copy the color or gray level of the source pixel - from its specific pixel coordinate - to the specified pixel coordinate in the destination image, the pixel locations corresponding destination coordinates are filled in with the colors corresponding to the *floor*, *ceil* or *round* integers linked to them - such as $dstCoord(floor)$ or $dstCoord(ceil)$ $dstCoord(round)$ - as shown in Table 2 and Figure 2. For example, in Table 2, if $dstCoord = 4$, $ceil = 3$.

Table 2: Nearest-neighbor algorithm scaling and rounding

| *dstCoord* | ***Equation 1*** | ***Calculated*** *scrCoord* | ***floor*** | ***ceil*** | ***round*** |
|---|---|---|---|---|---|
| 1 | 1 x (4/7) | 0.57 | 0 | 1 | 1 |
| 2 | 2 x (4/7) | 1.14 | 1 | 2 | 1 |
| 3 | 3 x (4/7) | 1.71 | 1 | 2 | 2 |
| 4 | 4 x (4/7) | 2.28 | 2 | 3 | 2 |
| 5 | 5 x (4/7) | 2.85 | 2 | 3 | 3 |
| 6 | 6 x (4/7) | 3.42 | 3 | 4 | 3 |
| 7 | 7 x (4/7) | 4 | 4 | 4 | 4 |

This means that the color to copy at the destination coordinate number 4 will be the same as that corresponding to the source coordinate 3, which is the blue color as shown in Figure 2's source image strip. In other words $4(3) = blue$.

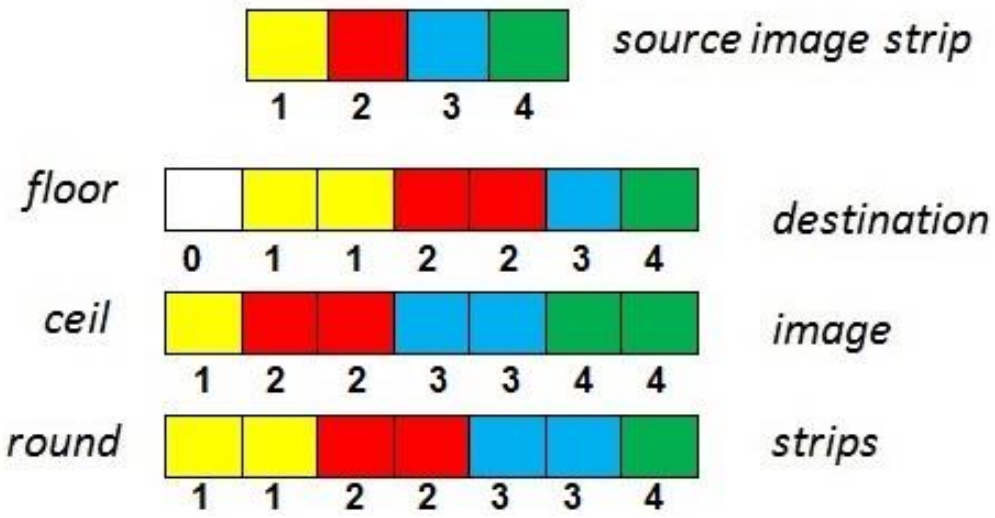

Figure 2: Image strip of length 4 upscaled to the length 7 following the nearest-neighbor interpolation

It is important to note that while using the *floor* function, one destination strip coordinate becomes invalid since there is no coordinate equal to zero number in the source image (this is only due to Matlab indexing). However, using the *ceil* function, all destination strip coordinates are valid and matched with their corresponding coordinates in the source image. The same when the round function is used. Again, it is important to note that, in all three cases presented, the gray levels or colors were copied differently due to different methods with different precisions for rounding-off purposes.

## 3. EXPERIMENTAL EVALUATIONS

The Matlab software is used as the simulation tool. Selected full- and no-reference Image Quality Assessment (IQA) metrics are namely Mean Squared Error (MSE), Blind/Referenceless Image Spatial Quality Evaluator (BRISQUE)[27], Naturalness Image Quality Evaluator (NIQE)[28] and Structural Similarity Index (SSIM)[29]. Inferential statistical analysis concept is applied to deduce from a small number of images and draw a conclusion of the behavior of each rounding function on a bigger number of images[24-26]. Since it would be unreasonable to seek to use the entire population of all existing standard test images, only ten sample or test images (shown in Figure 3) were carefully selected and downloaded from the USC-SIPI Image Database[30]. The scaling ratios varying from two to five were used. Thanks to the MS Picture Manager, images with the size 128 x 128 to 256 x 256, 170 x 170 to 510 x 510, 128 x 128 to 512 x 512 and 102 x 102 to 510 x 510 were downscaled and used. Note that IMAGE-9 (5.3.01) was cropped from its original image with 1024 x 1024 size using Matlab's *imcrop* function to get the face part and 512 x 512 image size wanted.

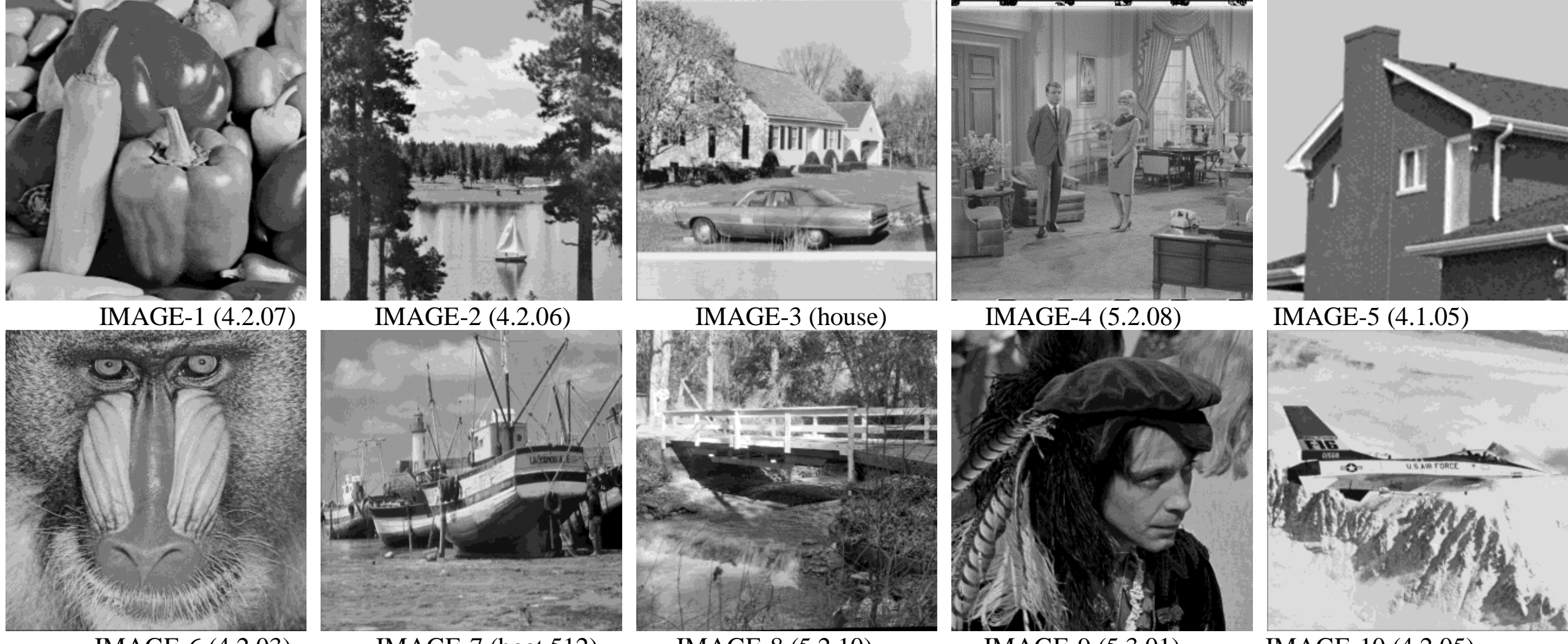

Figure 3: The above 512 x 512 images are found in the USC-SIPI Image Database[30] with the filenames between parentheses

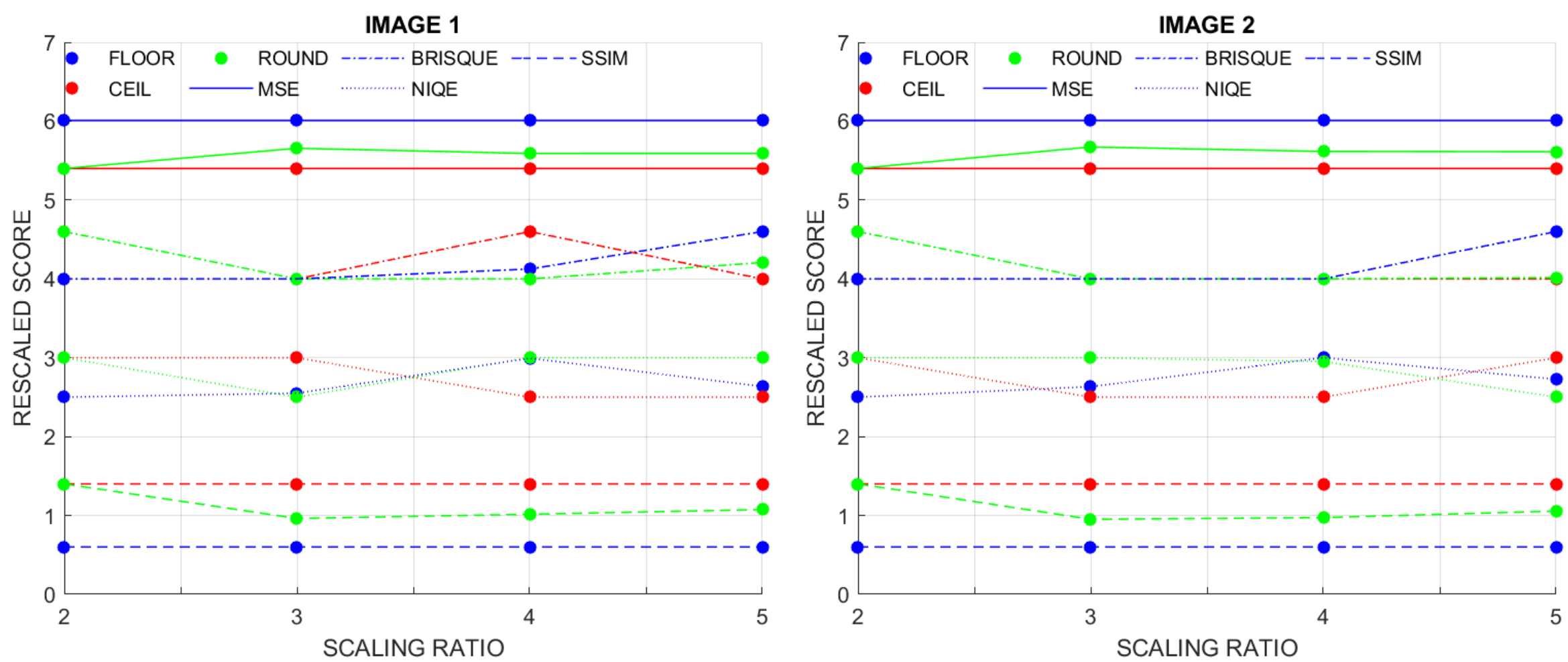

Figure 4: IMAGE-1 (4.2.07) and IMAGE-2 (4.2.06).

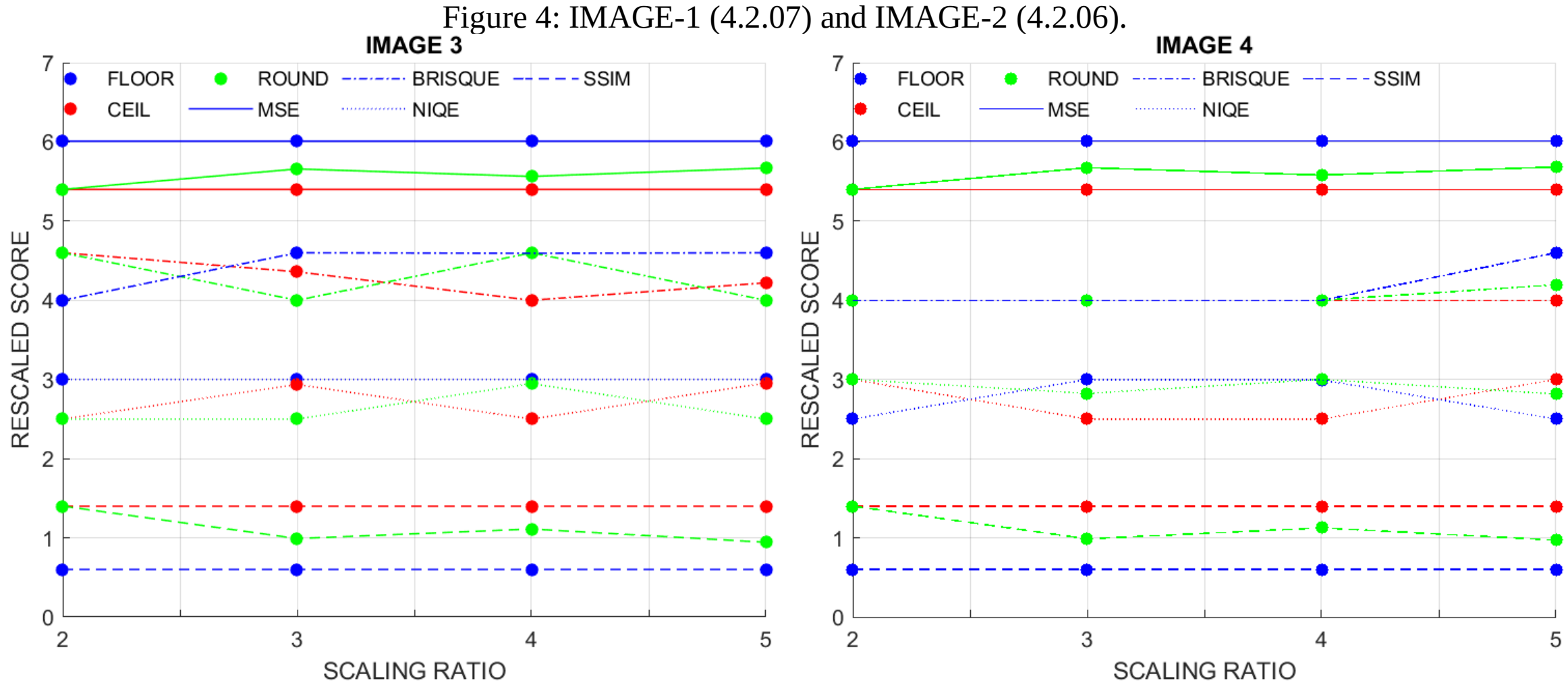

Figure 5: IMAGE-3 (house) and IMAGE-4 (5.2.08).

Figure 4, Figure 5, Figure 6, Figure 7 and Figure 8 show the graphical behaviors of rescaled SSIM, NIQE, BRISQUE, MSE scores to the intervals [0.6, 1.4], [2.5, 3.0], [4.0, 4.6], [5.4 and 6.01], respectively. Referring to the works done and presented in[18-20], the lower MSE, BRISQUE, and NIQE scores mean generally the better image quality while the higher SSIM score means generally the better image quality.

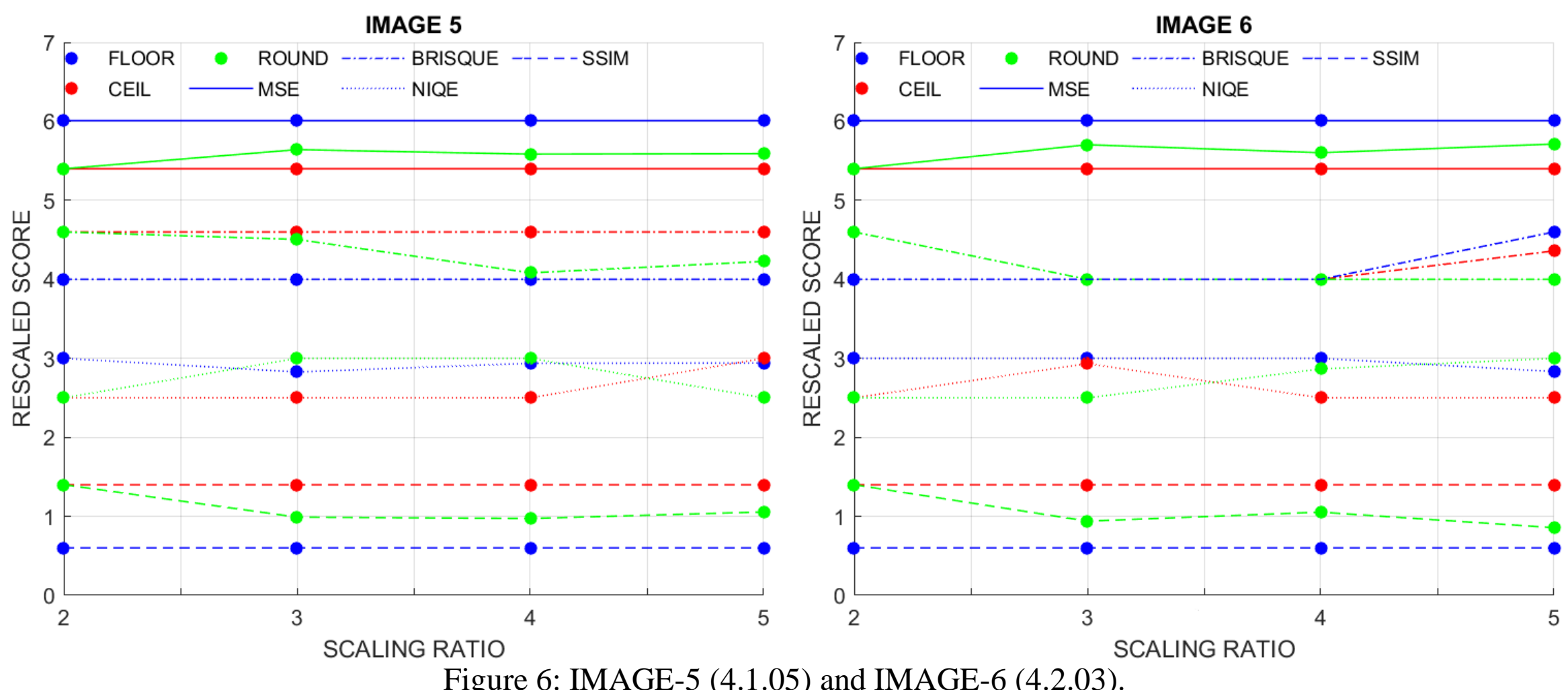

Figure 6: IMAGE-5 (4.1.05) and IMAGE-6 (4.2.03).

As can be seen in Figure 4, Figure 5, Figure 6, Figure 7 and Figure 8, it may seem unclear enough to readers which rounding function achieved the highest scores, due to the rescaling strategy adopted only for graphical representation purposes and whose rescaling intervals have been provided earlier in this part. Therefore, the highest scores achieved previously were collected from Matlab's data table and their corresponding rounding functions were provided, as C (i.e. *ceil function*), F (i.e. *floor function*) and R (i.e. *round function*), in Table 3, Table 4, Table 5, Table 6 and Table 7.

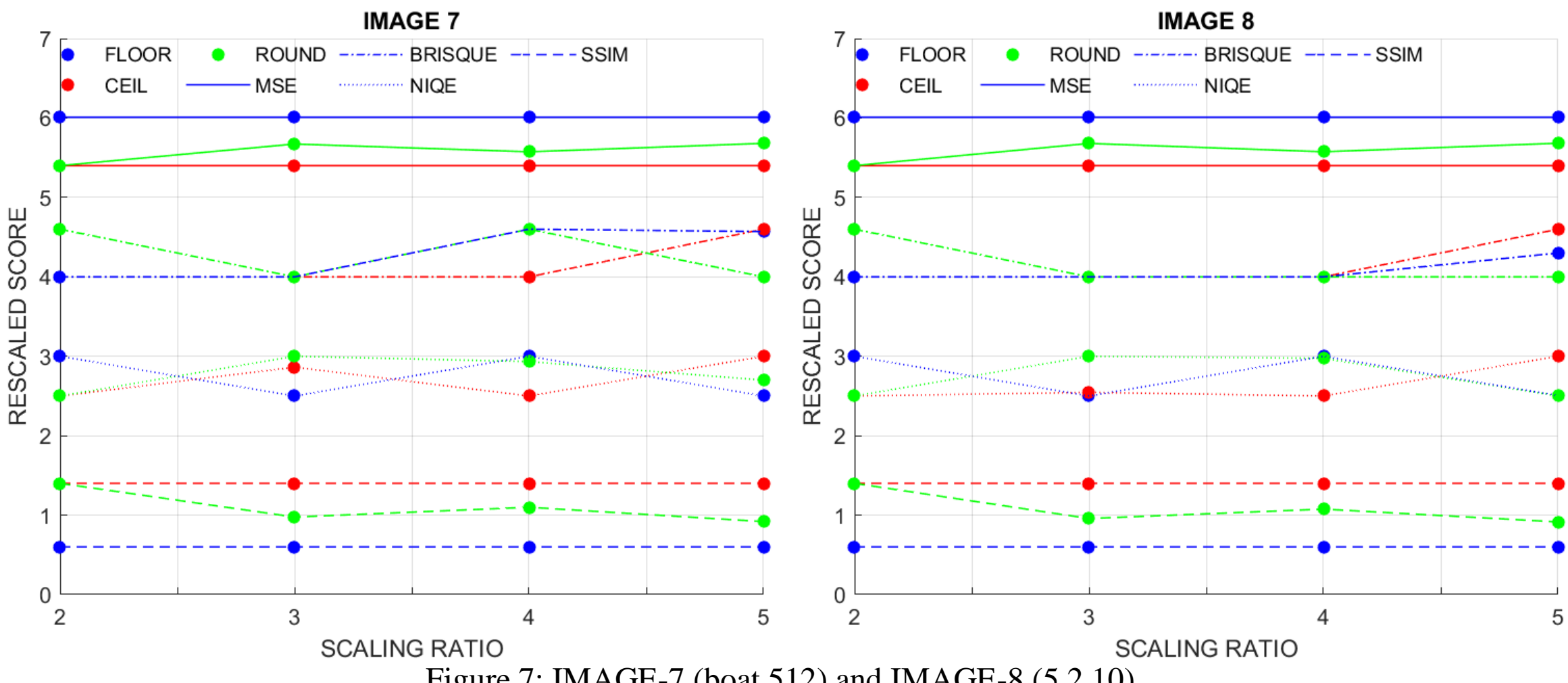

Figure 7: IMAGE-7 (boat.512) and IMAGE-8 (5.2.10).

Now, based on the number of test images, IQA metrics, and scaling ratios selected, each rounding function has a chance of achieving a maximum number of occurrences per image that is equal to 16. Here, the maximum number of occurrences is also referred to as the number of targeted occurrences. Such a number is directly proportional to the number of sample or test images, IQA metrics, and scaling ratios. Now, the best rounding function, for all cases examined, would ideally occur 16 times per image. However, that was not achieved because, as can be seen, not all rounding functions tied performances in every examined situation or case. Despite that, it can be seen immediately from these examples that the ceil function is repeated more times than other rounding functions. In other words, the number of occurrences of the *ceil* function looks to be much higher than that of the other two rounding functions. However, it is not enough to draw a conclusion based on observation, because, some may argue saying that this cannot be accurate enough to lead to an acceptable conclusion, especially when a smaller number of sample images was only used. Therefore, in the effort to alleviate such a concern, the inferential statistics concept is used to deduce from a small but representative sample the characteristics of a bigger population[24-26]. In other words, a small number of sample images can help to conclude what would be the behaviors of each rounding function within a bigger or very big number of images.

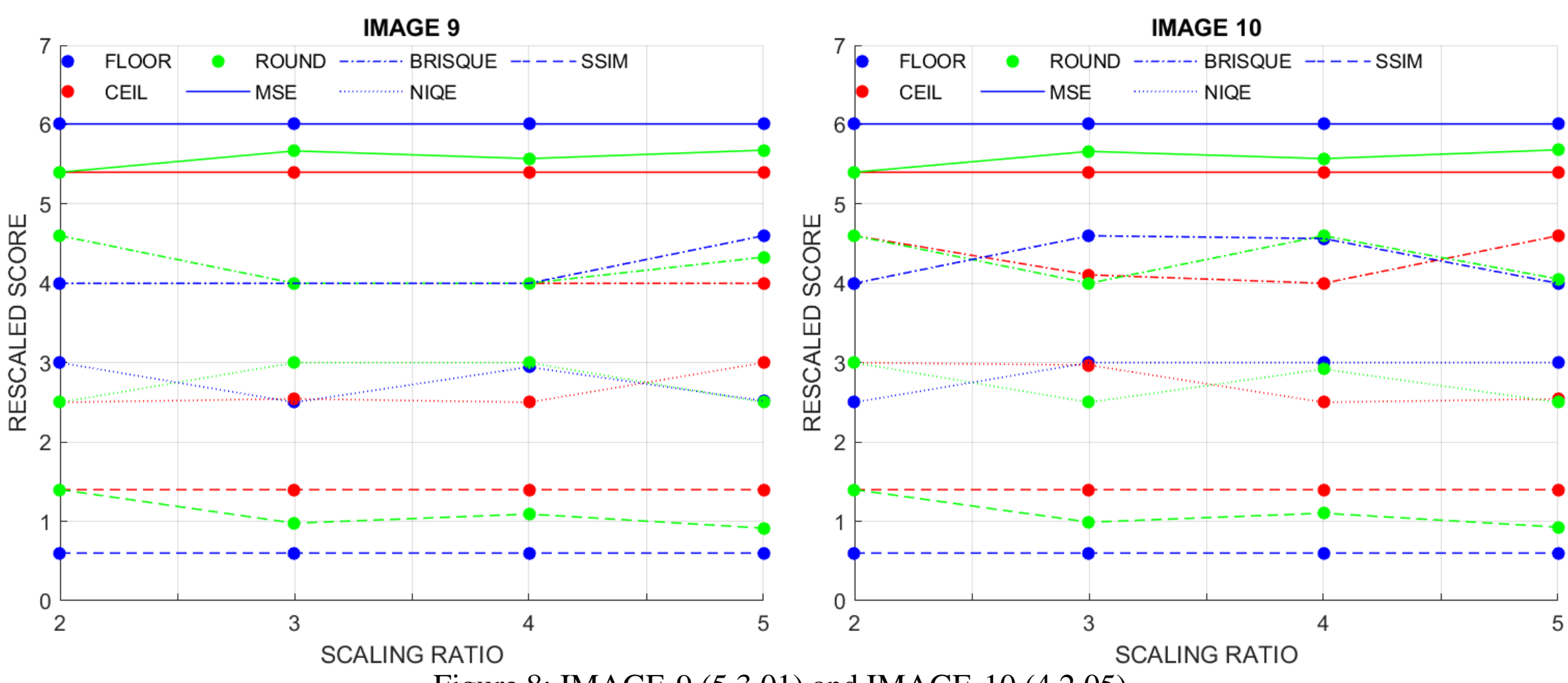

Figure 8: IMAGE-9 (5.3.01) and IMAGE-10 (4.2.05).

As mentioned earlier, each rounding function has a chance of achieving 16 targeted occurrences per image. This means that with all ten images presented/used in Figure 3, each rounding function has a chance of achieving 160 targeted occurrences. Table 8 presents the number of achieved occurrences over the number of targeted occurrences.

Table 3: F = 7 times, C = 27 times, R = 10 times

| **IMAGE 1 & 2** | ratio = 2 | ratio = 3 | ratio = 4 | ratio = 5 | ratio = 2 | ratio = 3 | ratio = 4 | ratio = 5 |
|---|---|---|---|---|---|---|---|---|
| MSE | C & R | C | C | C | C & R | C | C | C |
| BRISQUE | F & C | F & C & R | R | C | F & C | F & C & R | F & C & R | C |
| NIQE | F | R | C | C | F | C | C | R |
| SSIM | C & R | C | C | C | C & R | C | C | C |

Table 4: F = 6 times, C = 25 times, R = 12 times

| **IMAGE 3 & 4** | ratio = 2 | ratio = 3 | ratio = 4 | ratio = 5 | ratio = 2 | ratio = 3 | ratio = 4 | ratio = 5 |
|---|---|---|---|---|---|---|---|---|
| MSE | C & R | C | C | C | C & R | C | C | C |
| BRISQUE | F | R | C | R | F & C & R | F & C & R | F & C & R | C |
| NIQE | C & R | R | C | R | F | C | C | F |
| SSIM | C & R | C | C | C | C & R | C | C | C |

Table 5: F = 7 times, C = 24 times, R = 11 times

| **IMAGE 5 & 6** | ratio = 2 | ratio = 3 | ratio = 4 | ratio = 5 | ratio = 2 | ratio = 3 | ratio = 4 | ratio = 5 |
|---|---|---|---|---|---|---|---|---|
| MSE | C & R | C | C | C | C & R | C | C | C |
| BRISQUE | F | F | F | F | F | F & C & R | F & C & R | R |
| NIQE | C & R | C | C | R | C & R | R | C | C |
| SSIM | C & R | C | C | C | C & R | C | C | C |

As can be seen, none of the three rounding functions achieved the number of targeted occurrences, even on a small number of images equals to ten. Despite that, it is still important to seek to generalize how each rounding function would have performed if a bigger or very big number of images was used in the effort to alleviate the concern of just drawing a conclusion based on a small number of images. Let us consider 50 to 50 000 images. As can be seen in Table 9, such images correspond to targeted occurrences varying from 800 to 800 000. In Table 8, it is shown that with only 10 images, the *ceil* function achieves 78.75 % of 160 targeted occurrences. In other words, we are 95% confident that the *ceil* function can achieve 78.75 % of all targeted occurrences in 10 images presented in Figure 3. This means that, in 50 or more images, the ceil function can achieve at least 71 % of the corresponding targeted occurrences (i.e. 71 % = 78.75% - 7.75%). Note that the 95% rule and margin of error are widely explained in the statistics literature and are often used in inferential statistics[24-26, 31].

Table 6: F = 9 times, C = 26 times, R = 12 times

| **IMAGE 7 & 8** | ratio = 2 | ratio = 3 | ratio = 4 | ratio = 5 | ratio = 2 | ratio = 3 | ratio = 4 | ratio = 5 |
|---|---|---|---|---|---|---|---|---|
| MSE | C & R | C | C | C | C & R | C | C | C |
| BRISQUE | F & C | F & C & R | C | R | F & C | F & C & R | F & C & R | R |
| NIQE | C & R | F | C | F | C & R | F | C | F & R |
| SSIM | C & R | C | C | C | C & R | C | C | C |

Table 7: F = 7 times, C = 24 times, R = 11 times

| **IMAGE 9 & 10** | ratio = 2 | ratio = 3 | ratio = 4 | ratio = 5 | ratio = 2 | ratio = 3 | ratio = 4 | ratio = 5 |
|---|---|---|---|---|---|---|---|---|
| MSE | C & R | C | C | C | C & R | C | C | C |
| BRISQUE | F & C | F & C & R | F & C & R | C | F | R | C | F |
| NIQE | C & R | F | C | R | F | R | C | R |
| SSIM | C & R | C | C | C | C & R | C | C | C |

Table 8: Targeted occurrence number and percentage

| **Rounding function** | **Number of achieved occurrences/targeted** | **Percentage of achieved occurrences** |
|---|---|---|
| *ceil* (C) | 126/160 | 78.75% |
| *round* (R) | 56/160 | 35% |
| *floor* (F) | 36/160 | 22.5% |

Also, we are 95% confident that the *round* function can achieve 35 % of all targeted occurrences in 10 images in Figure 3, which means that, in 50 or more images, the *round* function can achieve at least 27.25 % of the corresponding targeted occurrences. Regarding the *floor* function, we are 95% confident that it can achieve 22.5 % of all targeted occurrences in those 10 images which means that, in 50 or more images, the *floor* function can achieve at least 14.75 % of the corresponding targeted occurrences. As can be seen in Table 9, note that the normal distribution is assumed and smaller the margin of error, the more confidence one may have that his/her results will be representative of the targeted number.

Table 9: Results obtained with a confidence level = 95%

| **A target number of images** | **A target number of occurrences** | **Margin of Error** |
|---|---|---|
| 50 | 800 | 6.93% |
| 500 | 8,000 | 7.67% |
| 5,000 | 80,000 | 7.74% |
| 50,000 | 800,000 | 7.75% |

## 4. CONCLUSION

In this study, we have reviewed and evaluated *ceil*, *floor*, and *round* functions using standard test images, IQA metrics, and inferential statistical analysis. Studying the influence of rounding functions on NN image interpolation encompasses the contribution and novelty of this paper. Demonstrations were provided to clarify how each rounding function works and influences NN image interpolation quality. The effects of loss of precision of rounding functions were examined in each rounding function case. The examination helped to determine the number of achieved and targeted occurrences. Inferential analysis-based experiments showed that with the level of confidence equals to 95%, the *ceil* function could achieve 78.75% of all targeted occurrences in 10 test images and minimum 71 % of the corresponding targeted occurrences in 50 or more images. Also, it was shown that the *round* and *floor* functions could achieve 35% and 22.5% of all targeted occurrences in 10 test images, respectively, as well as a respective minimum of 27.25 % and 14.75 % of the corresponding targeted occurrences in 50 or more images. It can be concluded that the *ceil* function is the most appropriate rounding function for nearest-neighbor interpolation purposes.

## CONFLICT OF INTEREST

On behalf of all authors, the corresponding author states that there is no conflict of interest.